# Deep Recurrent Factor Model: Interpretable Non-Linear and Time-Varying Multi-Factor Model

Kei Nakagawa[1] , Tomoki Ito[2], Masaya Abe[1] , Kiyoshi Izumi[2]
[1] Nomura Asset Management Co., Ltd., Tokyo, Japan
[2]School of Engineering, The University of Tokyo, Tokyo, Japan
{k-nakagawa, m-abe}@nomura-am.co.jp, m2015titoh@socsim.org, izumi@sys.t.u-tokyo.ac.jp

## Abstract

A linear multi-factor model is one of the most important tools in equity portfolio management. The linear multi-factor models are widely used because they can be easily interpreted. However, financial markets are not linear and their accuracy is limited. Recently, deep learning methods were proposed to predict stock return in terms of the multi-factor model. Although these methods perform quite well, they have significant disadvantages such as a lack of transparency and limitations in the interpretability of the prediction. It is thus difficult for institutional investors to use black-box-type machine learning techniques in actual investment practice because they should show accountability to their customers.

Consequently, the solution we propose is based on LSTM with LRP. Specifically, we extend the linear multi-factor model to be non-linear and time-varying with LSTM. Then, we approximate and linearize the learned LSTM models by LRP. We call this LSTM+LRP model a deep recurrent factor model. Finally, we perform an empirical analysis of the Japanese stock market and show that our recurrent model has better predictive capability than the traditional linear model and fully-connected deep learning methods.

## Introduction

Stock return has predictability been an important research theme, both academically and practically. In empirical finance, the typical method to predict stock returns is cross-section (regression) analysis using cross-sectional data of corporate attributes. The attribute that explains the stock return as revealed by the cross-section analysis is called a factor. A model that explains stock returns using multiple factors is called multi-factor model. A representative multi-factor model is the Fama–French three-factor model (Fama and French 1992; 1993), which proposed that the cross-sectional structure of the stock return can be explained by three factors: beta (market portfolio), size (market capitalization), value (PBR). Since then, many other factors were identified. Specifically, Harvey, Liu, and Zhu (2016) reported over 300 identified factors by 2012.

Almost all these multi-factor models, including the Fama-French three-factor model, are linear regression models and until now, linear multi-factor models are one of the most important tools in equity portfolio management. Institutional investors are accountable to those whose money they invest. Linear multi-factor models are so widely used because they can be easily interpreted. There are two ways to use a linear multi-factor model: as a return or risk model, where the return model is used for predicting stock returns and the risk model for describing the factor attribution of the predicted stock return. However, it is well-known that financial markets are not linear (Mandelbrot 1997).

Recently, deep learning methods were proposed to predict stock returns in terms of multi-factor models (Levin 1996; Abe and Nakayama 2018). Deep learning has been proven to be a powerful machine learning technique with various applications (LeCun, Bengio, and Hinton 2015), and can take the nonlinearity of factors into consideration. However, it has not yet made possible the consideration of time-varying factors. Therefore, we propose implementing a non-linear time-varying multi-factor model with long short-term memory (LSTM), which is a representative model to express a time-varying non-linear relationship. Although LSTM performs well in many sequential prediction problems, it has significant disadvantages, such as a lack of transparency and limitations in the interpretability of the prediction. As it is difficult for institutional investors to use black-box-type machine learning technique such as LSTM in actual investment practice because they need to be accountable to their customers, we present the application of layer-wise relevance propagation (LRP) (Bach et al. 2015) to linearize the proposed LSTM model. We consider this LSTM+LRP model a deep recurrent factor model. We can model non-linear and time-varying factors as a return model and comprehend which factor contributes to the prediction as a risk model.

The remainder of this paper is organized as follows. We formulate our deep recurrent factor model in the next section. Then, we empirically analyze the Japanese stock market and show that our model has a better predictive capability than the traditional linear model and fully-connected deep learning methods. Finally, we review related works and conclude.

# Deep Recurrent Factor Model with LSTM-LRP

## Non-Linear and Time-Varying Multi-Factor Model

Here, we formulate the deep recurrent factor model. The traditional linear multi-factor model assumes that stock return $r_i$ can be described as follows:

$$r_i = \alpha_i + X_{i1}F_1 + \cdots + X_{iN}F_N + \varepsilon_i, \quad (1)$$

where $F_n$ are a set of factor returns, $X_{in}$ denotes the $i$-th stock's exposure to factor $n$, $\alpha_i$ is an intercept term assumed to be equal to a risk-free rate of return under the arbitrage pricing theory framework (Ross 2013). $\varepsilon_i$ is a random term with mean 0 and assumed to be uncorrelated across other random terms. Usually, factor exposure $X_{in}$ is defined by the linearity of several descriptors.

While linear multi-factor models are effective tools for equity portfolio management, the assumption of a linear relationship is rather restrictive. Specifically, linear models assume that each factor affects the stock return independently, but ignore the possible interactions between factors and the time-dependency of factor exposures and returns as shown in Table 1.

We extend equation (1) to the non-linear time-dependent model:

$$\begin{aligned} r_i(t) &= \tilde{f}(X_{i1}(t)F_1(t) + X_{i1}(t-1)F_1(t-1) + \\ & \quad \cdots + X_{iN}(t)F_N(t) + \cdots) + \varepsilon_i(t), \end{aligned} \quad (2)$$

where $r_i(t), \varepsilon_i(t), F_n(t), X_{in}(t)$ are respectively $i$'s stock return, random term, $n$-th factor return, and factor exposure at time $t$. Here, $\tilde{f}$ is a non-linear function that does not satisfy either or both of the following conditions:

$$f(x+y) = f(x) + f(y), \quad (3)$$
$$f(\alpha x) = \alpha f(x). \quad (4)$$

Prediction with the non-linear time-varying model (2) is more complex than that with the linear one since it requires both factor returns and unknown function $\tilde{f}$. Similar to a previous study (Levin 1996), we assume factor returns are constants $\bar{F}_n$. This assumption is also used in a linear multi-factor model. The prediction task can thus be simplified, since the factor returns are no longer variables.

Time-dependent model (2) can be transformed as follows:

$$\begin{aligned} r_i(t) &= \tilde{f}(X_{i1}(t)F_1(t) + X_{i1}(t-1)F_1(t-1) + \\ & \quad \cdots + X_{iN}(t)F_N(t) + \cdots) + \varepsilon_i(t) \\ &= f(X_{i1}(t), X_{i1}(t-1), \ldots, X_{iN}(t), \ldots) + \varepsilon_i(t) \end{aligned} \quad (5)$$

To estimate the unknown non-linear function $f$, we use LSTM model.

## Implementation with LSTM-LRP

We use the LSTM model (Hochreiter and Schmidhuber 1997; Gers and Schmidhuber 2000) as a representation of equation (14):

$$\begin{aligned} i_t &= \sigma(W_i h_{t-1} + U_i x_t + b_i) & (6) \\ f_t &= \sigma(W_f h_{t-1} + U_f x_t + b_f) & (7) \\ o_t &= \sigma(W_o h_{t-1} + U_o x_t + b_o) & (8) \\ g_t &= \tanh(W_g h_{t-1} + U_g x_t + b_g) & (9) \\ c_t &= f_t \odot c_{t-1} + i_t \odot g_t & (10) \\ h_t &= o_t \odot \tanh(c_t), & (11) \end{aligned}$$

where $\sigma$ is the sigmoid function and $i, f, o, c$ are, respectively, the input, forget, output gate, and cell memory values in response to input vector at time $t$. Here, the initial values are $c_0 = 0$ and $h_0 = 0$. Above, the activation functions $\sigma$ and $\tanh$ are applied element-wise. Moreover, operator $\odot$ denotes the Hadamard product(element-wise product).

Next, from the viewpoint of the risk model, we calculate the decomposition of the predicted return by LRP. LRP is a method used to explain the output of the neural network based on input values. Specifically, LRP distributes the output to the input through relevance score $R$. The relevance score is propagated from the output to the input layer in a form close to a chain rule.

Two types of relevance calculation methods, weighted and multiplicative connections, are proposed to calculate the LRP of the LSTM network (Arras et al. 2017).

**Weighted connections** Assume we know relevance scores $R_j$ of neurons $z_j$ and want to compute scores $R_i$ of lower-layer neurons $z_i$. Values $z_j$ are computed during forward-pass as follows:

$$z_j = \sum_i (w_{ij} \times z_i + b_j). \quad (12)$$

Then, relevance scores $R_{i \leftarrow j}$ are calculated as:

$$R_{i \leftarrow j} = \frac{w_{ij} \times z_i + \frac{\varepsilon \times \text{sign}(z_j + b_j)}{N}}{z_j + \varepsilon \times \text{sign}(z_j)} \times R_j, \quad (13)$$

where $N$ is the number of lower-layer neurons to which $z_j$ is connected and $\varepsilon$ is a small positive real number to ensure non-negativity.

**Multiplicative connections** To apply LRP to LSTM layers, we have to consider the gating mechanism. Let $z_j$ of the upper-layer be calculated as $z_j = z_g \times z_s$ by lower layers $z_g$ and $z_s$. Here, $z_g$ is the gate, whose activation is between 0 and 1, and $z_s$ is the source that carries the information from the lower layer or previous layers. Then, the backpropagation rule is simply $R_g = 0$ and $R_s = R_j$. It may seem that the values of $z_g$ and $z_s$ are disregarded; however, they were already considered in $R_j$, which depends on $z_j$.

We linearize the LSTM models by LRP and rewrite the relevance score $R_j$ corresponding to input $X_{in}(t)$ with $\beta_{in}(t)$:

$$\begin{aligned} r_i(t) &= f(X_{i1}(t), X_{i1}(t-1), \ldots, X_{iN}(t), \ldots) + \varepsilon_i(t) \\ &\approx \beta_{i1}(t) X_{i1}(t) + \ldots + \beta_{iN} X_{iN}(t), \ldots + \varepsilon_i(t). (14) \end{aligned}$$

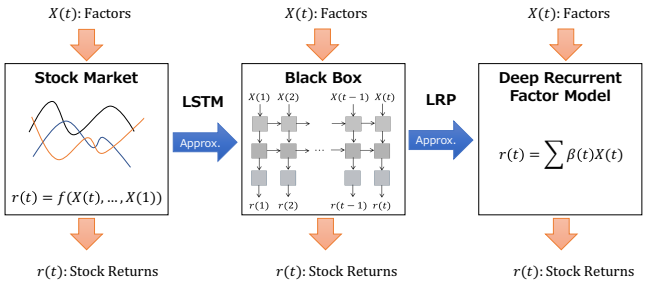

Figure 1: Deep recurrent factor model

Here, $\beta_{i1}(t)$ can be interpreted as a factor return linearized by LRP. The factor return of the traditional linear multi-factor model is calculated for the investment universe, whereas the factor return linearized by LRP is calculated for each stocks in the investment universe. We can model the non-linearity and time-dependency of factors as the return model and identify which factor contributes to the prediction as a risk model. Figure 1 shows the outline of the deep recurrent factor model.

## Experiment on Japanese Stock Markets

### Data

We prepare a dataset of TOPIX500 index constituents. TOPIX500 is a well-accepted stock market index for the Tokyo Stock Exchange (TSE) in Japan and comprises the large and mid-cap segments of the Japanese market. The index is also often used as a benchmark for overseas institutional investors investing in Japanese stocks.

We use the five factors and 16 factor descriptors listed in Table 1. These are used relatively often in practice and are studied most widely in academia (Jurczenko 2015).

In calculating these factors, we acquire the necessary data from the Nikkei Portfolio Master (NPM) and Bloomberg. Factor descriptors are calculated on a monthly basis (at the end of month) from December 1990 to March 2015 as input data and stock returns with dividends are acquired on a monthly basis (at the end of month) as output data.

### Models

Our problem is identifying prediction model $f(x)$ of an output $Y$, for the next month's stock returns given an input $X$ and various descriptors. One set of training data is shown in Table 2. In addition to the proposed deep recurrent factor model, we use a linear regression model as baseline, support vector regression (SVR(Drucker et al. 1997)), random forest(Breiman 2001), and fully-connected deep learning (deep factor model) as comparison methods. The deep factor model and deep recurrent factor model are implemented with Chainer (Tokui et al. 2015) and the comparison methods with scikit-learn (Pedregosa et al. 2011). Table 3 lists the details of each model.

We train all models by using the latest 60 sets of training data from the past five years. We decided the hyper parameters of deep factor and deep recurrent factor models using 1/60 of each training dataset as the validation dataset. The models are updated by sliding one month ahead and carrying out a monthly forecast. The prediction period is from April 2007 to March 2015. To verify the effectiveness of each method, we compare the prediction accuracies of these models and the profitability of the quintile portfolio. We also construct a long/short portfolio strategy for a net-zero investment to buy top stocks and sell bottom stocks with equal weights in quintile portfolios. For quintile portfolio performance, we calculate the annualized average return, volatility, and Sharpe ratio(= return/volatility). Additionally, we calculate the average mean absolute error (MAE) and root mean squared error (RMSE) for the prediction period as prediction accuracy.

### Results

Table 4 lists the average MAEs and RMSEs of all years, along with the annualized return, volatility, and Sharpe ratio for each method. The best values appears in bold on each row.

The LSTM model has the best prediction accuracy in terms of MAE and RMSE. On the other hand, the LSTM+LRP model is the most profitable in terms of the Sharpe ratio. In any case, we find that the LSTM and DNN models exceed the linear model in terms of accuracy and profitability. This implies that the relationship between stock returns on the financial market and the factor is non-linear. We also find that the LSTM model exceeds the DNN model and the LSTM+LRP model exceeds the DNN+LRP model in terms of accuracy and profitability. These facts imply that the relationship between stock returns on the financial market and the factor is time-varying. Although the accuracy decreased slightly because of the linearization by LRP, the deep recurrent factor model, which can capture such a non-linear and time-varying relationship, is thought to be superior.

### Interpretation

Here, we confirm that the effect of the LRP approximation on performance is high in terms of the return model. We interpret the top quintile portfolio based on the factor using the linear, deep factor, and deep recurrent factor models as of March 2015.

The contributions of each descriptor calculated by LRP are summed up for each factor and displayed as percentiles. We can thus identify which factor contributes to the prediction as a risk model compared to linear model, as shown in Figure 2.

Although the values of the contributions are different in each model, the order is almost the same. Value, size, quality, risk, and momentum factors contribute to the prediction in this order. As described above, the contribution is similar to the linear model, but prediction performance is better than the linear model, as shown in Table 4. We can observe that the size and value factors account for more than half of the contribution to the top quintile portfolio. Generally, the momentum factor is not very effective, but the value factor is effective on the Japanese stock market (Fama and French 2012). For practitioners, we feel this is in line with the actual Japanese stock market.

Table 1: Factors and factor descriptors

| Factors | Descriptors | Formulas |
|---|---|---|
| Risk | 60VOL | Standard deviation of stock returns in the past 60 months (Ang et al. 2006) |
| | BETA | Regression coefficient of stock returns and market risk premium (Sharpe 1964) |
| | SKEW | Skewness of stock returns in the past 60 months (Boyer, Mitton, and Vorkink 2009) |
| Quality | ROE | Net income/Net assets (Soliman 2008) |
| | ROA | Operating Profit/Total assets (Soliman 2008) |
| | ACCRUALS | Operating cashflow Operating profit (Novy-Marx 2013) |
| | LEVERAGE | Total liabilities/Total assets (Soliman 2008) |
| Momentum | 12-1MOM | Stock returns in the past 12 months except for the last month (Jegadeesh and Titman 1993) |
| | 1MOM | Stock returns in the past month (Jegadeesh and Titman 1993) |
| | 60MOM | Stock returns in the past 60 months (Jegadeesh and Titman 2001) |
| Value | PSR | Sales/Market value (Suzuki 1998) |
| | PER | Net income/Market value (Basu 1977) |
| | PBR | Net assets/Market value (Fama and French 1992) |
| | PCFR | Operating cashflow/Market value (Hou, Karolyi, and Kho 2011) |
| Size | CAP | log(Market value) (Fama and French 1992) |
| | ILLIQ | average(abs(Stock returns)/Trading volume) (Amihud 2002) |

Table 2: One set of training data for March 2015

| Input: 80 dim | Output: 1 dim |
|---|---|
| Factor descriptors: 16 × 5 dim | Return: 1 dim |
| February 2015 November 2014 August 2014 May 2014 February 2014 | March 2015 |

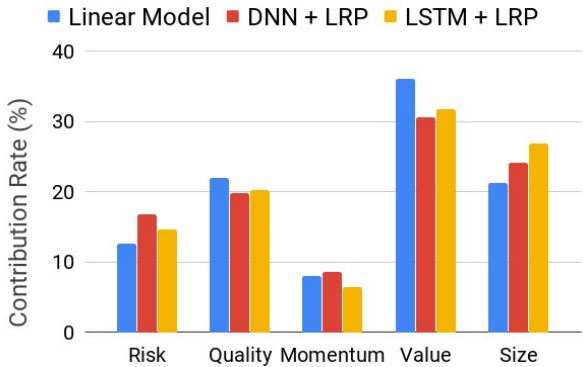

Figure 2: Factor contributions to the prediction by linear, deep factor, and deep recurrent factor models

## Related Works

Many studies on stock return predictability using machine learning have been published. Cavalcante et al. (2016) presented a review of the application of several machine learning methods in finance. In their survey, most of these were forecasts of stock market returns. However, there is no paper that deals with the prediction method in terms of a multi-factor model.

Levin (1996) discussed the use of multilayer feed forward neural networks for predicting stock returns within the framework of the multi-factor model. Abe and Nakayama (2018) extended this model to deep learning and investigated the performance of the method on the Japanese stock market. They showed that deep neural networks generally outperform shallow ones, and the best networks also outperform representative machine learning models. These results indicate deep learning holds promise as a skillful machine learning method to predict stock returns in cross-section.

However, these works are only for use as a return model, and the problem is that the viewpoint of a risk model is lacking. Nakagawa, Uchida, and Aoshima (2018) proposed the application of LRP to decompose the attributes of the predicted return as a risk model. However, they do not examine the influence on performance due to the approximation of LRP and not considering the time-dependency of factors. We thus extend this model to a time-varying multi-factor model with LSTM+LRP and analyze the effects of the LRP approximation on performance on the Japanese stock market.

## Conclusions

We proposed a deep recurrent factor model that is nonlinear and time-varying multi-factor model implemented with LSTM+LRP. The empirical analysis of the Japanese stock market shows the LRP approximation is effective in terms of return and risk models. Our model can capture nonlinear and time-varying relationship with factors and stock return in an interpretable way.

In terms of further study, we would like to confirm the effectiveness of our model on stock markets other than the Japanese one. Although we considered 16 factors, some other macroeconomic variables, such as foreign exchange rates, interest rates, and consumer price index, can be used as inputs.

Table 3: Details on each method

| Model | | Description |
|---|---|---|
| Deep recurrent factor model | LSTM | We used one forward and one backward LSTM cell, where the dimensions of the hidden layers and sequence length in the LSTMs are 16 and 5, respectively. The epoch was 10 with early stopping, which was decided using 1/60 of each training dataset as validation dataset. We used Adam (Kinga and Adam 2015) for the optimization algorithm. |
| | LSTM+LRP | We used one forward and one backward LSTM cell, with the same setting as the LSTM and the LRP method with bias factor = 0.0, as described in Arras et al. (2017). |
| Deep factor model | DNN | We used the multi-layer perceptron model, where the dimensions of the hidden layers were 40 or 80, number of the hidden layers between 1 and 5, and epoch 10 with early stopping. We decided on the hyper parameters in the same manner as with the LSTM. We used ReLU as the activation function and Adam (Kinga and Adam 2015) for the optimization algorithm. |
| | DNN+LRP | We used the multi-layer perceptron model with the same settings as the DNN, and the LRP method with bias factor = 0.0 as described in Arras et al. (2017). |
| Linear model | | Linear model was implemented with scikit-learn with class "sklearn.linear_model.LinearRegression." All parameters were default values in this class. |
| Support Vector Regression | | Support vector regression (SVR) was implemented with scikit-learn with class "sklearn.svm.SVR." All parameters were default values in this class. |
| Random Forest | | Random Forest was implemented with scikit-learn with class "sklearn.ensemble.RandomForestRegressor." All parameters were default values in this class. |

Table 4: Average MAEs and RMSEs of all years and annualized returns, volatilities, and Sharpe ratios for each method

| | Deep recurrent factor model | | Deep factor model | | Linear model | SVR | Random forest |
|---|---|---|---|---|---|---|---|
| | LSTM | LSTM+LRP | DNN | DNN+LRP | | | |
| Return[%] | 13.20 | **15.18** | 11.23 | 11.33 | 6.02 | 6.63 | 5.44 |
| Volatility [%] | 15.01 | **11.60** | 14.90 | 14.77 | 13.31 | 17.52 | 11.69 |
| Sharpe Ratio | 0.88 | **1.31** | 0.75 | 0.77 | 0.45 | 0.38 | 0.47 |
| MAE | **0.07492** | 0.07654 | 0.07570 | 0.07671 | 0.08033 | 0.08497 | 0.08231 |
| RMSE | **0.09830** | 0.09912 | 0.09914 | 0.09999 | 0.10533 | 0.10911 | 0.10791 |

# Acknowledgements
We thank Tsuyoshi Ogawa and Yuya Morooka for discussions and insights that helped clarify the ideas in this paper.